\documentclass[letterpaper, 10 pt, conference]{ieeeconf}  

\IEEEoverridecommandlockouts                              
\usepackage{comment}
\usepackage{mathbbol}
\usepackage{tabularx}
\usepackage{xcolor, soul}
\sethlcolor{green}

\let\labelindent\relax

\usepackage{hyperref}    
\hypersetup{
    colorlinks=true,      
    linkcolor=blue,       
    citecolor=blue,       
    filecolor=magenta,    
    urlcolor=cyan         
}

\usepackage{etoolbox}
\makeatletter
\patchcmd{\@makecaption}
  {\scshape}
  {}
  {}
  {}
\makeatother
\usepackage[caption=false]{subfig}

\usepackage{algorithmic}
\usepackage{euscript}[mathcal]
\usepackage[inline]{enumitem}
\usepackage{multirow}
\usepackage{amsmath}
\usepackage{esvect}
\usepackage{color}

\usepackage{url}

\definecolor{newcolor}{rgb}{0.858, 0.188, 0.478}
\usepackage[english]{babel}
\usepackage[caption=false]{subfig}
\usepackage{multirow}
\usepackage{wrapfig}
\usepackage{booktabs}
\usepackage{array}

\usepackage{stfloats}
\usepackage[utf8]{inputenc}
\usepackage[T1]{fontenc}
\usepackage{amsmath}

\usepackage{graphicx}
\usepackage{adjustbox}
\graphicspath{{./images/}}
\usepackage{amssymb}
\usepackage{longtable}

\makeatletter 
\@namedef{ver@float.sty}{3000/12/31}
\makeatother

\usepackage{mathptmx} 

\title{\LARGE \bf
An AutoML-based approach for Network Intrusion Detection}

\author{Nana Kankam Gyimah$^{*}$, 
Judith Mwakalonge$^{*}$, Gurcan Comert$^{+}$, 
Saidi Siuhi$^{*}$, \\
Robert Akinie$^{+}$, Methusela Sulle$^{*}$, Denis Ruganuza$^{*}$, Benibo Izison$^{*}$, Arthur Mukwaya$^{*}$ \\
$^{*}$ South Carolina State University, Orangeburg, South Carolina, US, 29117\\ 
$^{+}$ North Carolina A\&T State University, Greensboro, North Carolina, US, 27411\\  
}

\begin{document}

\maketitle
\thispagestyle{empty}
 \pagestyle{empty}

\begin{abstract}
In this paper, we present an automated machine learning (AutoML) approach for network intrusion detection, leveraging a stacked ensemble model developed using the MLJAR AutoML framework. Our methodology combines multiple machine learning algorithms, including LightGBM, CatBoost, and XGBoost, to enhance detection accuracy and robustness. By automating model selection, feature engineering, and hyperparameter tuning, our approach reduces the manual overhead typically associated with traditional machine learning methods. Extensive experimentation on the NSL-KDD dataset demonstrates that the stacked ensemble model outperforms individual models, achieving high accuracy and minimizing false positives. Our findings underscore the benefits of using AutoML for network intrusion detection, as the AutoML-driven stacked ensemble achieved the highest performance with 90\% accuracy and an 89\% F1 score, outperforming individual models like Random Forest (78\% accuracy, 78\% F1 score), XGBoost and CatBoost (both 80\% accuracy, 80\% F1 score), and LightGBM (78\% accuracy, 78\% F1 score), providing a more adaptable and efficient solution for network security applications.

Index Terms - Network Intrusion Detection, AutoML, Stacked Ensemble Model, Cybersecurity, NSL-KDD Dataset
\end{abstract}

\section{Introduction}
Network Intrusion Detection (NID) is a crucial aspect of cybersecurity, aimed at identifying unauthorized access or malicious activity within a network \cite{chaabouni2019network}. Its importance has grown significantly as networks expand and become increasingly complex, with the potential consequences of security breaches becoming more severe. According to the Special Report on Cyberwarfare In The C-Suite, Cybersecurity Ventures projects that global cybercrime costs will grow by 15 percent per year over the next five years, reaching $\$10.5$ trillion USD annually by 2025, up from $\$3$ trillion USD in 2015 \cite{cybersecurityventuresCybercrimeCost}. This surge marks the greatest transfer of economic wealth in history, threatening innovation and investment, exceeding annual damages from natural disasters, and surpassing the profits of the global illegal drug trade.

Effective NID systems are essential in mitigating these risks by providing early detection of intrusions and facilitating rapid response \cite{rawindaran2021cost}. With the average time to identify and contain a breach estimated at 277 days, early detection is crucial in reducing both the duration and impact of security incidents \cite{ibmCostData}. Consequently, the role of effective NID systems is increasingly vital in safeguarding sensitive data, maintaining business continuity, and protecting organizational assets from cyber threats.

In tackling these threats, network intrusion detection primarily relies on two methodologies: signature-based detection and machine learning-based detection \cite{zhang2022comparative}. Signature-based detection remains widely adopted due to its high accuracy in identifying known threats by matching network activity against a predefined database of malicious signatures \cite{ioulianou2018signature}. However, this method has several notable drawbacks:
\begin{enumerate}
    \item Reliance on known patterns of malicious activity, making them ineffective against zero-day attacks and emerging threats that have not yet been cataloged.
    
    \item High maintenance requirements arise because signature databases need frequent updates to remain effective against evolving attack patterns, making the process both resource-intensive and time-consuming.
  
    \item Performance issues with large signature databases arise as the database of known signatures expands over time, leading to detection delays and negatively impacting overall system efficiency.    
\end{enumerate}

Addressing the limitations of signature-based detection methods, machine learning (ML) methods \cite{panda2011network, li2014new} are increasingly favored in network intrusion detection systems. Unlike the classical methods that require constant updates, ML-based detection can analyze vast amounts of network data to identify complex and nuanced patterns associated with both known and previously unknown malicious behavior \cite{habeeb2022network}. Additionally, this adaptability is particularly advantageous given the rapidly changing nature of cyber threats, as seen in cases like the NotPetya attack \cite{wiredUntoldStory}, where ML systems were able to detect anomalous patterns in data traffic before the attack was fully understood \cite{salem2024advancing}.

\begin{figure*}[h!]
\centering
\includegraphics[scale=0.8]{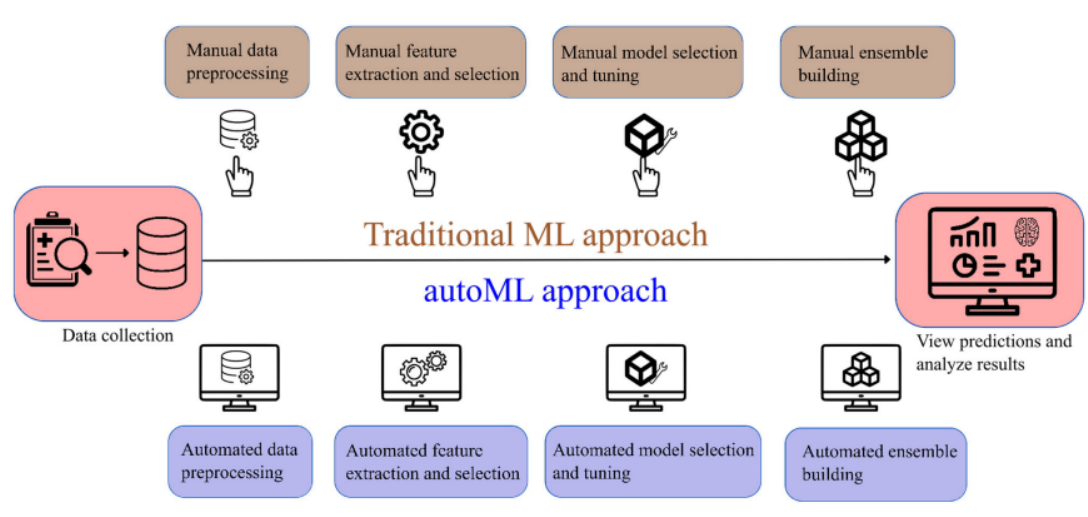}
\caption{A comparative overview of traditional ML and AutoML approaches. Traditional ML requires manual steps like data preprocessing, feature extraction, model selection, and tuning, while AutoML automates these tasks, streamlining the workflow from data collection to predictions \cite{raj2023automl}.}
\label{fig:Trad-ML vs AutoML}
\end{figure*}

Traditional machine learning (ML) methods, however, still face certain limitations as they rely heavily depend on manual processes, including feature engineering; feature extraction, and selection \cite{wagh2013survey, garg2017novel} as shown in Fig. \ref{fig:Trad-ML vs AutoML}. These processes, along with tasks like model selection and tuning add further complexity, making it challenging to handle large-scale intrusion data efficiently. As a result, these techniques often suffer from reduced recognition accuracy and higher false alarm rates.

To overcome the limitations of traditional machine learning, Automated Machine Learning (AutoML) approaches \cite{salehin2024automl} have emerged as powerful alternatives by automating the ML workflow, from data preprocessing to model validation \cite{hutter2019automated}, as shown in Fig. \ref{fig:Trad-ML vs AutoML} eliminating the need for manual feature selection, enhancing performance, and adapting to evolving data characteristics. In this paper, we adopt the MLJAR AutoML framework \cite{mljar} to develop AutoML-based intrusion detection algorithms that improve recognition accuracy, and reduce false alarm rates. The key contributions of this work are as follows:

\begin{enumerate}
    \item Development of AutoML-based prediction algorithms for network intrusion detection to enhance detection accuracy, minimize false positives, and improve overall system robustness. 
    
    \item A comparative study on the adaptation of conventional ML methods and AutoML-based prediction is undertaken to evaluate their effectiveness in addressing the network intrusion detection tasks. 
          
    \item Extensive experiments on the NSL-KDD dataset demonstrate the effectiveness of the developed AutoML-based algorithms in boosting detection accuracy and reducing false alarm rates, outperforming traditional ML-based approaches.
\end{enumerate}

The rest of this paper is structured as follows: Section $\text{II}$ provides an overview of related research on ML-based intrusion detection approaches. Section $\text{III}$ outlines the methodology adopted in this study. Section $\text{IV}$ covers the dataset, experimental setup, hyperparameter configurations, and evaluation metrics. Section $\text{V}$ discusses the experimental results and insights gained. Finally, Section $\text{VI}$ concludes the paper and outlines potential future research directions.

\section{Related Works}
This section reviews relevant literature on ML-based intrusion detection methods. As cyber threats evolve, traditional signature-based techniques often fail to detect new attacks. Machine learning (ML) offers a more adaptive solution by analyzing network traffic patterns to identify anomalies. This review highlights recent advancements and their impact on network intrusion detection.

In \cite{bitton2019machine}, a network anomaly detection system was proposed for remote desktop connections using machine learning algorithms. An exponential random graph model was introduced in \cite{tsikerdekis2021network} to integrate network topology statistics for accurate anomaly detection in network domains. In \cite{peng2018detection}, an SDN-based flow prediction method using a double P-value with the K-NN algorithm was proposed, demonstrating higher precision, lower false positives, and significantly better adaptation than other models. Similarly, in \cite{ma2021novel}, a linear kernel-based SVM model was used for anomaly detection in traffic profiles, employing natural language processing techniques to preprocess data and extract feature vectors before classification. In \cite{camacho2019group}, an anomaly detection model extending PCA with group-wise PCA and additional exploratory features was developed based on recent field applications. Additionally, a novel lightweight framework \cite{roy2022lightweight} was developed to predict anomalies and cyber-attacks in IoT networks, with evaluations on the NSL-KDD and CICIDS2017 datasets showing improved detection rates and lower false alarm rates. Lastly, \cite{li2022dfaid} proposed an active anomaly detection approach based on deviation- and density-aware features to address abnormal traffic, with a focus on enhancing data privacy.

\section{Proposed Methodology}
The proposed methodology establishes an AutoML framework \cite{mljar}, as illustrated in Fig. \ref{fig:Trad-ML vs AutoML}, to improve network intrusion detection accuracy by automating model selection, feature engineering, and optimization, thus enhancing detection efficiency and reducing false positives. Serving as the foundation for model development and selection, this framework streamlines the entire machine learning process—from feature engineering and hyperparameter tuning to model validation—resulting in highly effective models that fully leverage the training data. Additionally, its automated retraining and hyperparameter tuning make it well-suited for dynamic tasks like network intrusion detection, where evolving traffic patterns require adaptive models. Performance is further boosted through ensemble learning and stacking techniques, while automated feature selection optimizes detection speed and accuracy.

The adoption of the MLJAR AutoML framework in our study facilitated the development of an ensemble-stacked model specifically tailored for network intrusion detection. This robust model leverages a combination of algorithms—namely, LightGBM, CatBoost, and XGBoost variants—assigned with strategically calibrated weights to optimize accuracy. By integrating these models, the ensemble maximizes the strengths of each algorithm, enhancing predictive performance, stability, and adaptability to varied data patterns. This approach results in a balanced and resilient model, providing significant improvements in both detection accuracy and overall robustness.



To extensively compare traditional ML models with the proposed AutoML framework, we evaluate these models across several dimensions, including accuracy, precision, recall, and the $F1$-score. By examining key metrics such as prediction accuracy, and the outlined metrics, this study aims to highlight the practical advantages and limitations of each approach, offering valuable insights into their suitability for various machine learning tasks.

\subsubsection{\textbf{Random Forest}}
This is an ensemble learning method that improves predictive accuracy by combining multiple decision trees. It constructs a "forest" of trees by selecting random subsets of features and data samples, mitigating overfitting and variance. At each node, the algorithm evaluates a subset of features to find the best split, minimizing an impurity measure like Gini impurity (Eq. \ref{eq:gini}):

\begin{align}
    G_{\text{impurity}} = 1 - \sum_{i=1}^{c} p_i^2,
    \label{eq:gini}
\end{align}

where \( p_i \) is the proportion of class \( i \) in the node, and \( c \) is the number of classes.

Final predictions for instance \( x_i \) are made by aggregating the predictions of all trees via majority voting (Eq. \ref{eq:rf-prediction}):

\begin{align}
    \hat{y}_i = \text{majority\_vote}(T_1(x_i), T_2(x_i), \ldots, T_K(x_i)),
    \label{eq:rf-prediction}
\end{align}

Random Forest is effective due to its robustness to overfitting, ability to handle high-dimensional data, and interpretability via feature importance analysis.

\subsubsection{\textbf{XGBoost (Extreme Gradient Boosting)}}
This implementation of the gradient boosting algorithm is both efficient and scalable, featuring regularization, handling of missing data, and parallel computation. The model iteratively builds an ensemble of decision trees, minimizing a regularized objective function (Eq. \ref{eq:xgboost}):

\begin{align}
    \mathcal{L}(\theta) = \sum_{i=1}^{n} l(y_i, \hat{y}_i) + \sum_{k=1}^{K} \Omega(f_k),
    \label{eq:xgboost}
\end{align}

where \( l(y_i, \hat{y}_i) \) is the loss function and \( \Omega(f_k) \) penalizes model complexity.

At each iteration, a new tree corrects residuals, updating predictions:

\[
    \hat{y}_i^{(t)} = \hat{y}_i^{(t-1)} + f_t(x_i)
\]

XGBoost leverages first-order gradients \( g_i \) and second-order Hessians \( h_i \) for precise optimization. The regularization term (Eq. \ref{eq:xgboost-reg}) helps control overfitting:

\begin{align}
    \Omega(f_k) = \gamma T + \frac{1}{2} \lambda \|\omega\|^2,
    \label{eq:xgboost-reg}
\end{align}

This, combined with parallelization, makes XGBoost highly effective for large-scale predictive modeling tasks.

\subsubsection{\textbf{CatBoost}}
This is a gradient boosting algorithm optimized for handling categorical features and designed for high efficiency in both accuracy and speed. Unlike traditional methods, it uses innovative encoding techniques for categorical variables, ideal for datasets with high cardinality. The regularized objective function (Eq. \ref{eq:catboost}) is minimized as follows:

\begin{align}
    \mathcal{L}(\theta) = \sum_{i=1}^{n} l(y_i, \hat{y}_i) + \sum_{k=1}^{K} \Omega(f_k),
    \label{eq:catboost}
\end{align}

where \( l(y_i, \hat{y}_i) \) is the loss function, and \( \Omega(f_k) \) regularizes model complexity.

At each iteration, a new tree minimizes residuals: \( \hat{y}_i^{(t)} = \hat{y}_i^{(t-1)} + f_t(x_i) \). CatBoost uses Ordered Boosting to reduce prediction shift and Symmetric Trees to improve both accuracy and speed. The regularization term (Eq. \ref{eq:catboost-reg}) helps control overfitting:

\begin{align}
    \Omega(f_k) = \lambda \|\omega\|^2 + \alpha T,
    \label{eq:catboost-reg}
\end{align}

These features, combined with gradient and second-order gradient information, make CatBoost highly effective for large-scale, high-dimensional datasets with categorical variables.

\subsubsection{\textbf{LightGBM(LGBM)}}
This is a highly efficient gradient boosting implementation, designed for improved speed and scalability. Unlike traditional methods, LGBM uses a leaf-wise tree growth strategy, splitting the leaf with the maximum loss reduction, which results in deeper trees and higher accuracy. It is well-suited for large datasets due to its ability to handle sparse features, reduce memory usage, and support parallel learning.

The objective function minimizes a regularized loss, as shown in Eq. \ref{eq: lgbm}:

\begin{align}
    \mathcal{L}(\theta) = \sum_{i=1}^{n} l(y_i, \hat{y}_i) + \sum_{k=1}^{K} \Omega(f_k),
    \label{eq: lgbm}
\end{align}

where \( l(y_i, \hat{y}_i) \) is the loss function and \( \Omega(f_k) \) controls model complexity.

At each iteration, LGBM adds a new tree based on leaf-wise splits, updating predictions:

\[
    \hat{y}_i^{(t)} = \hat{y}_i^{(t-1)} + f_t(x_i)
\]

LGBM uses first- and second-order gradients for efficient optimization. The regularization function (Eq. \ref{eq:lgbm-reg}) is:

\begin{align}
    \Omega(f_k) = \gamma T + \frac{1}{2} \lambda \|\omega\|^2,
    \label{eq:lgbm-reg}
\end{align}

This controls overfitting while ensuring performance. With its ability to handle large data and provide accurate predictions, LGBM is widely adopted for tasks requiring speed and memory efficiency.

\section{Dataset, Experiment Setup, Hyper-parameter tuning and Evaluation Metrics}
This section describes the dataset, experimental setup, hyperparameter tuning, and performance evaluation metrics used in the network intrusion detection models.

\subsection{Dataset}
\label{subsection:Dataset}
The dataset used in this study is the NSL-KDD dataset \cite{dhanabal2015study}, a refined version of the widely recognized KDD Cup 1999 dataset \cite{engen2011exploring}. The NSL-KDD dataset is a popular benchmark for evaluating network intrusion detection (NID) systems. Unlike its predecessor, the NSL-KDD addresses issues with redundant records and provides a more balanced distribution of data across both the training and testing sets, which improves the fairness and robustness of system evaluations. Specifically, it includes 125,973 records for training (NSL-KDDTrain+) and 22,544 records for testing (NSL-KDDTest+), ensuring comprehensive testing across diverse traffic patterns as shown in Table \ref{tab:class_distribution}.

\begin{table}[h!]
\caption{Distribution of classes in the NSL-KDD training and testing sets.}
\centering 
\begin{tabular}{c c c c c c c}
\hline
         & \textbf{normal} & \textbf{Dos} & \textbf{Probe} & \textbf{U2R} & \textbf{R2L} & \textbf{Total} \\ \hline
\textbf{Train} & 67,343          & 45,927        & 11,656         & 52          & 995          & 125,973       \\
\textbf{Test}  & 9,711           & 7,458         & 2,421          & 200         & 2,754        & 22,543        \\ \bottomrule
\end{tabular}
\label{tab:class_distribution}
\end{table}

The dataset consists of five classes: Normal, Denial of Service (DoS), User to Root (U2R), Remote to Local (R2L), and Probe, each representing different types of network traffic as shown in Table \ref{tab:class_distribution}. Normal corresponds to legitimate traffic, while the four attack classes reflect various malicious activities. The dataset includes 41 features, divided into three categories: basic features (1–10), such as protocol type and service; content features (11–22), such as the number of failed login attempts; and traffic features (23–41), which include the number of connections to the same host within a given time window \cite{dhanabal2015study}.

\subsection{Hyper-parameter Tuning}
The hyperparameters of the classifiers used in the NID models are optimized using Bayesian optimization with 10-fold cross-validation. In this process, the dataset is split into 10 equally sized subsets (folds), with each algorithm trained on 9 folds and validated on the remaining one. This is repeated 10 times, allowing each fold to serve as the validation set once. The average performance across all folds ensures robust model evaluation, minimizing reliance on any specific train-test split. The data is shuffled before splitting to diversify training and validation sets, reducing overfitting and improving generalization. Stratification is applied to maintain class proportions in each fold. The best-performing configuration is selected based on average performance across the folds, as summarized in Table \ref{table:validation accuracy}.

\begin{table}[h!]
\caption{Classifiers Hyper-parameter Settings.}  
\centering 
\footnotesize
{\begin{tabular}{c c c} 
\hline
Classifier & Hyper-parameter & Values\\ [0.3ex]
\hline

& Criterion & Gini \\
& max\_features & 0.5 \\
Random Forest & min\_samples\_split & 20 \\
& max\_depth & 4 \\
\hline

& learning\_rate (eta) & 0.075 \\
& max\_depth & 8 \\
XGBoost & min\_child\_weight & 5 \\
& subsample & 1.0 \\
& colsample\_bytree & 1.0 \\
\hline

& learning\_rate & 0.05 \\
& depth & 8 \\
CatBoost & rsm & 0.8 \\
\hline

& num\_leaves & 63 \\
& learning\_rate & 0.05 \\
LightGBM & feature\_fraction & 0.9 \\
& bagging\_fraction & 0.9 \\
& min\_data\_in\_leaf & 5 \\
\hline 
\end{tabular}}
\label{table:validation accuracy}
\end{table}

\subsection{Evaluation Metrics}
The weighted average values of Precision, Recall, and F1-score, as defined in Eqs. \ref{eq:precision} through \ref{eq:F1-score}, along with accuracy as defined in Eq. \ref{eq:accuracy}, are adopted to evaluate the performance of the NID models \cite{al2023machine}. These metrics are calculated using the true positive ($tp_{i}$), true negative ($tn_{i}$), false positive ($fp_{i}$), and false negative ($fn_{i}$) values for each class $C_{i}$, where $i = 1, \cdots, m$ and $m$ represents the total number of classes in the dataset. Here, $|Y_{i}|$ denotes the total number of samples assigned to each class.

\begin{align} 
\text{Accuracy} = \frac{TP + TN}{TP + TN + FP + FN} 
\label{eq:accuracy} 
\end{align}

\begin{align}
Weighted\ Average\ Precision =\frac{\sum_{i=1}^{m}|Y_{i}| \frac{tp_{i}}{tp_{i} + fp_{i}}}{\sum_{i}^{m}|Y_{i}|} 
\label{eq:precision}
\end{align}

\begin{align}
Weighted\ Average\ Recall = \frac{\sum_{i=1}^{m}|Y_{i}| \frac{tp_{i}}{tp_{i} + fn_{i}}}{\sum_{i}^{m}|Y_{i}|} 
\label{eq:Recall}
\end{align}

\begin{align}
Weighted\ Average\ F1-score = \frac{\sum_{i=1}^{m}|y_{i}| \frac{2tp_{i}}{2tp_{i} + fp_{i} + fn_{i}}}{\sum_{i}^{m}|y_{i}|}  
\label{eq:F1-score}
\end{align}

\subsection{Experiment Setup}
The dataset used for the experiments is described in section \ref{subsection:Dataset}. The training set (NSL-KDDTrain+) which includes 125,973 records is used to train the intrusion detection models  and the testing dataset (NSL-KDDTest+) which includes 22,544 records is used to evaluate the detection models. For this study, the dataset has been transformed into binary classification tasks, distinguishing between Normal and Attack (malicious) traffic where the the four attack classes reflect various malicious activities.

To evaluate the performance of the proposed framework, Random Forest (RF), Xgboost, Catboost, lgbm and the adopted AutoML framework classifiers are used to evaluate the performance of the NID models. An $NVIDIA-SMI$ $A100-SXM4$-$40GB$ GPU on $Google Colab+$ is used in training the models.

\section{Results and Discussion}
Table \ref{tab:combined_results} demonstrate the performance metrics; Accuracy, Precision, Recall, and $F1$ Score of different machine learning models, including Random Forest, XGBoost, CatBoost, and LightGBM, as well as the final Stacked Ensemble model, for network intrusion detection.

\begin{table}[h!]
\centering
\caption{Performance metrics for evaluated models, with precision, recall, and $F1-score$ reported as weighted averages.}
\begin{tabular}{@{}lcccc@{}}
\toprule
\textbf{Model}   & \textbf{Accuracy} & \textbf{Precision} & \textbf{Recall} & \textbf{F1 Score} \\ \midrule
Random Forest         & 78   & 84   & 78   & 78 \\
XGBoost          & 80   & 85  & 80   & 80 \\
CatBoost         & 80   & 85  & 80   & 80 \\ 
LGBM          & 78   & 84   &  78  & 78 \\
\textbf{Stacked Ensemble} & \textbf{90}   & \textbf{90}   & \textbf{89}   & \textbf{89} \\
\bottomrule
\end{tabular}
\label{tab:combined_results}
\end{table}

\begin{figure}[h!]
\centering
\includegraphics[scale=0.34]{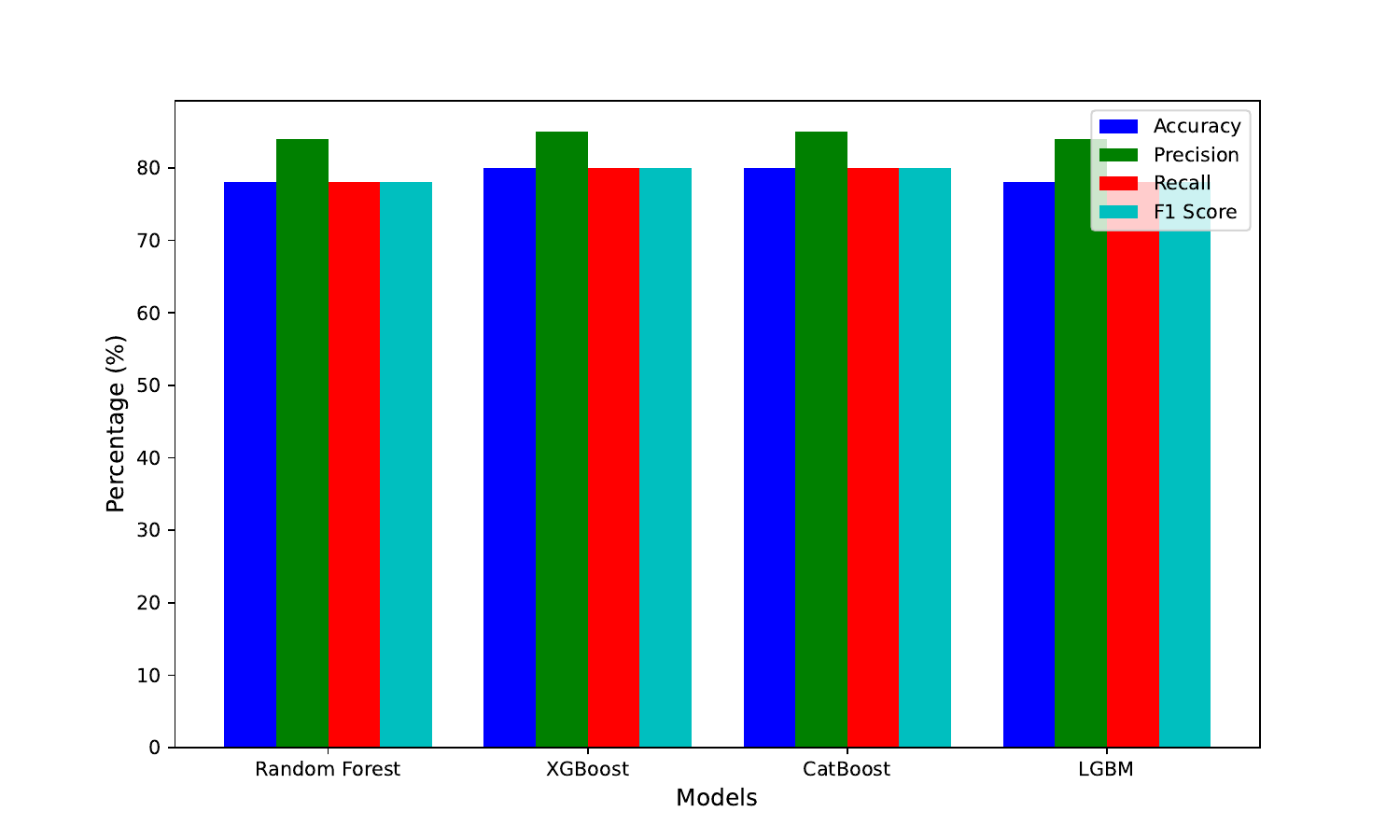}
\caption{Comparison of performance metrics (Accuracy, Precision, Recall, and F1 Score) for individual models (Random Forest, XGBoost, CatBoost, and LGBM) in network intrusion detection, highlighting variations across key metrics.}
\label{fig:model evaluation}
\end{figure}

Among the individual models as shown in Fg. \ref{fig:model evaluation}, XGBoost and CatBoost show the highest accuracy $(80\%)$ and maintain precision, recall, and F1 scores at 85 and 80, respectively, indicating solid overall performance. LightGBM and Random Forest both have slightly lower accuracy $(78\%)$ and are relatively less effective in comparison, with Precision, Recall, and F1 scores at $84\%$ and $78\%$, showing less balance in capturing true positives without increasing false positives.

\begin{figure}[h!]
\centering
\includegraphics[scale=0.3]{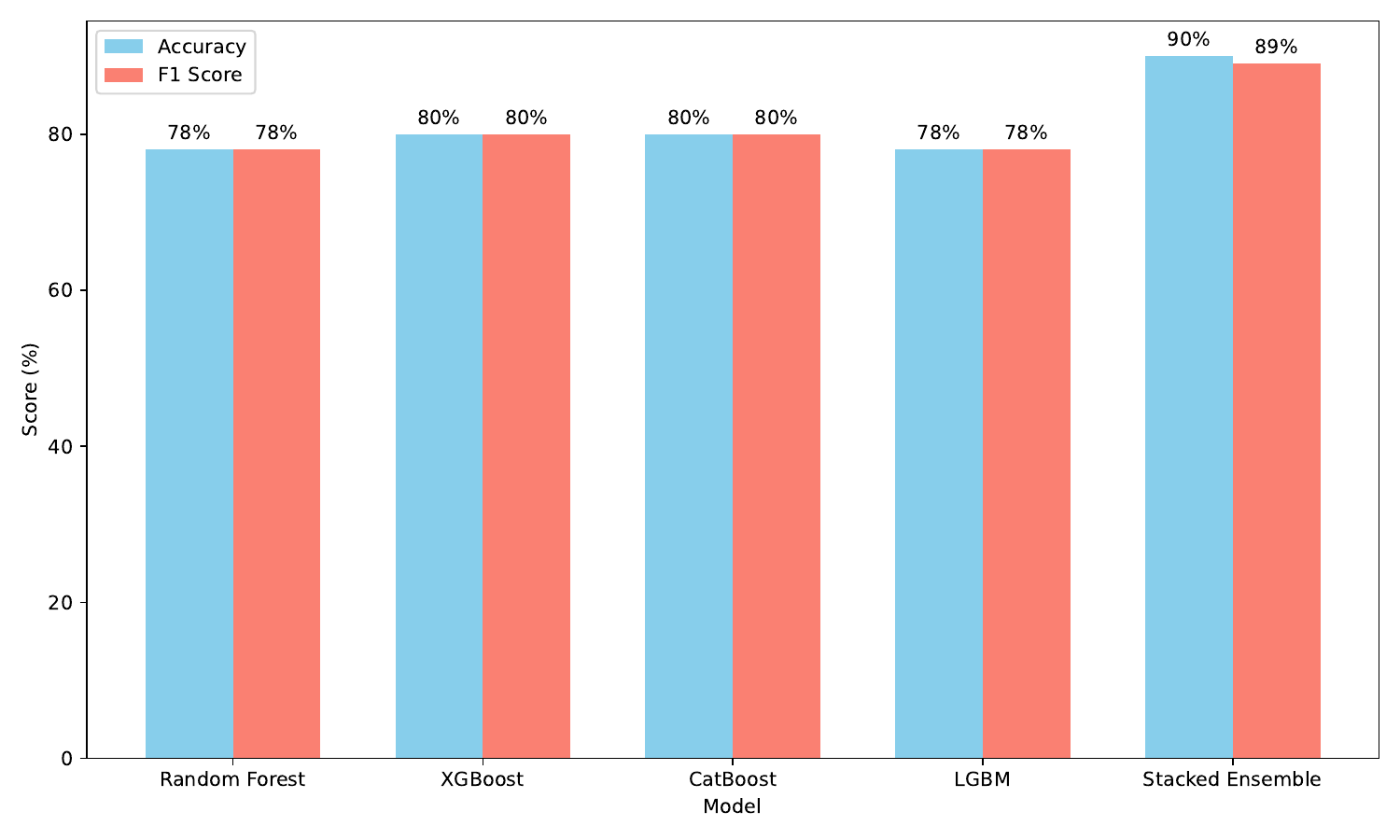}
\caption{Comparison of accuracy and F1 score across individual models (Random Forest, XGBoost, CatBoost, and LGBM) and the AutoML Stacked Ensemble model. The Stacked Ensemble achieves the highest performance, demonstrating the effectiveness of ensemble stacking for network intrusion detection.}
\label{fig:single_vs_autoMl}
\end{figure}

The Stacked Ensemble model, however, outperforms all individual models across all metrics as shown in Table \ref{tab:combined_results} and Fig. \ref{fig:single_vs_autoMl}. It achieves the highest accuracy at $90\%$, along with Precision, Recall, and F1 scores of $90\%$, $89\%$, and $89\%$, respectively. This improvement highlights the advantage of the AutoML ensemble-stacked approach, which combines the strengths of the individual models and ensemble models to create a more robust and reliable predictive model. The high precision and recall indicate that the Stacked Ensemble model is not only more accurate but also effective in minimizing false positives and capturing true positives, essential for tasks like network intrusion detection where precision and adaptability are critical.

\section{Conclusion and Future Work}
In this paper, we presented the development and evaluation of a Stacked Ensemble model for network intrusion detection, which was generated through an AutoML framework. This AutoML approach gave rise to the Stacked Ensemble model by combining the predictive strengths of diverse algorithms—Random Forest, XGBoost, CatBoost, and LightGBM—each contributing unique advantages in precision, recall, and adaptability, while addressing individual model limitations, such as overfitting or sensitivity to certain data patterns. The resulting stacked structure delivers a balanced, robust, and reliable detection system. Additionally, the AutoML-driven ensemble stacking approach not only enhances overall performance metrics but also streamlines the machine learning workflow by automating model selection, feature engineering, and hyperparameter tuning. This seamless integration allows the model to adapt to dynamic environments, where evolving network traffic patterns and emerging attack types demand responsive and flexible solutions. Such adaptability is critical for real-time intrusion detection, where high accuracy and prompt detection are essential for maintaining system security. 

Our future work will investigate the integration of Explainable AI (XAI) to identify top-performing features, enhancing model interpretability in network intrusion detection. We also plan to apply this approach to a large-scale surface defect dataset with multiple defect classes to develop more robust, generalized models. This focus on XAI and diverse datasets aims to improve both model transparency and adaptability for real-world applications.

\bibliographystyle{IEEEtran}
\bibliography{sample}

\end{document}